\documentclass[final]{l4dc2024}

\title[Information and Semi-Markov Decision Processes]
{Increasing Information for Model Predictive Control
with Semi-Markov Decision Processes}
\usepackage{times}

\usepackage{amsmath}
\usepackage{amssymb}
\usepackage{bbm}
\usepackage{mathtools}
\usepackage{csquotes}
\usepackage{wrapfig}
\usepackage{algpseudocode}
\usepackage{algorithm}
\usepackage{footmisc}
\usepackage[font=small]{caption}

\coltauthor{
    \Name{Rémy Hosseinkhan-Boucher}$^*$
    \Email{remy.hosseinkhan@universite-paris-saclay.fr}\\
    \Name{Stella Douka}$^*$
    \Email{styliani.douka@universite-paris-saclay.fr}\\
    \Name{Onofrio Semeraro}
    \Email{onofrio.semeraro@universite-paris-saclay.fr}\\
    \Name{Lionel Mathelin}
    \Email{lionel.mathelin@lisn.upsaclay.fr}\\
    \\
    \addr Université Paris-Saclay, Orsay, France \\
    CNRS,
    Laboratoire Interdisciplinaire des Sciences du Numérique,
    Orsay, France
}

\newcommand{\customfootnotetext}[2]{
    \begingroup
    \renewcommand{\thefootnote}{#1}
    \footnotetext{#2}
    \endgroup
}

\DeclareMathOperator*{\argmax}{arg\,max}
\DeclareMathOperator*{\argmin}{arg\,min}
\DeclareMathOperator*{\support}{Supp}

\newcommand{\stateProcess}{X}
\newcommand{\controlProcess}{U}
\newcommand{\historyProcess}{H}
\newcommand{\disturbanceProcess}{\eta}

\newcommand{\dynamicsOperator}{F}
\newcommand{\objectiveFunciton}{J}
\newcommand{\probabilityOperator}{P}
\newcommand{\stateTransitionProbability}{\mathcal{P}}

\newcommand{\policy}{\pi}
\newcommand{\MPCpolicy}{\pi^{\text{MPC}}}
\newcommand{\MPCpolicyEstimator}{\widehat{\pi}^{\text{MPC}}}

\newcommand{\costFunction}{c}

\newcommand{\timeHorizon}{T}

\newcommand{\startEquilibrium}{x_e}
\newcommand{\statePoint}{x}
\newcommand{\controlPoint}{u}
\newcommand{\disturbancePoint}{z}
\newcommand{\timeDelayPoint}{t}
\newcommand{\startEquilibriumStd}{\sigma_e}

\newcommand{\expectedInformationGain}{\text{EIG}}

\newcommand{\entropy}{\mathcal{H}}

\newcommand{\discreteTimeIndex}{k}

\newcommand{\samplingIteration}{n}
\newcommand{\randomDiscreteTime}{\kappa}
\newcommand{\randomInterdecisionTime}{\tau}

\newcommand{\SMTIP}{\text{SM-TIP}}
\newcommand{\TIP}{\text{TIP}}
\newcommand{\stateControlDataset}{\mathcal{D}}

\newcommand{\GPmean}{\mu}
\newcommand{\GPcovariance}{\Sigma}

\newcommand{\timehorizonMPC}{T^{\text{MPC}}}

\newcommand{\stateSpace}{\mathcal{X}}
\newcommand{\controlSpace}{\mathcal{U}}
\newcommand{\disturbanceSpace}{\mathcal{Z}}
\newcommand{\sampleSpace}{\Omega}
\newcommand{\sampleSpaceSigmaAlgebra}{\mathcal{F}}
\newcommand{\timeDelaySpace}{\mathcal{T}}

\begin{document}

    \maketitle
    
    \begin{abstract}%
        Recent works in Learning-Based Model Predictive Control of
        dynamical systems show impressive sample complexity performances
        using criteria from Information Theory to accelerate
        the learning procedure.
        However, the sequential exploration opportunities are limited by
        the system local state, restraining the amount of information
        of the observations from the current exploration trajectory.
        This article resolves this limitation by introducing temporal abstraction through the framework
        of Semi-Markov Decision Processes.
        The framework increases the
        total information of the gathered data for a fixed sampling budget, thus reducing the sample complexity.

    \end{abstract}


    \section[Introduction]{Introduction}\label{sec:introduction}

    \textit{Machine Learning Control (MLC)} is an interdisciplinary area
    of statistical learning and control theory which solves model-free
    optimal control problems~\citep{duriez2016}.
    Among the multiple approaches of the vast field of data-driven control,
    two classes have received notable attention by the machine learning
    community:
    \textit{Learning-Based Model Predictive Control (LB-MPC)}~\citep{hewing2020}
    and \textit{Model-Based Reinforcement Learning (MB-RL)}
    ~\citep{abbeel2006, recht2018, moerland2022}.
    The former refers to the combination of
    \textit{Model Predictive Control (MPC)}, an optimisation method based on
    a sufficiently descriptive model of the system dynamics~\citep{grune2011},
    and learning methods which enable the improvement of
    the prediction model from recorded data while possibly modeling
    uncertainty~\citep{aswani2013, koller2019}.
    The latter combines general function approximators such as
    linear models~\citep{tsitsiklis1997}, or more generally
    neural networks~\citep{sutton1999}, with
    \emph{Dynamic Programming (DP)}~\citep{bellman1957} principles to solve the underlying optimisation problem.

    \customfootnotetext{*}{Both authors contributed equally to this work.}
    {\setlength{\parskip}{1pt}
    Despite the recent impressive results in learning complex
    dynamical models~\citep{ha2018}, the \emph{sample complexity} of
    the learning process remains a major issue in the field of
    data-driven control~\citep[and see the references therein]{kakade2003, li2021}, in which
    the sample complexity is defined as the sample size required
    to learn a good approximation of the target concept~\citep{mohri2018}.
    For this reason, recent works~\citep{mehta2022, mehta2022b} in LB-MPC
    have focused on the design of exploration strategies based on the
    Information Theory concept of \emph{Expected Information Gain (EIG)}
    or negative \emph{Conditional Mutual Information (CMI)}~\citep{lindley1956}.
    The resulting criterion allows for quantifying the gain of information given by a new state-control observation on the estimated optimal system trajectory.
    Hence, this tool can be used as an acquisition function to guide the exploration of the state-control space.
    The concept of acquisition function is borrowed from the field of \emph{Bayesian Optimisation (BO)}.
    In particular, the work of~\cite{mehta2022} relies on the broader black-box BO framework of~\cite{neiswanger2021}.

    In a setting where the data is collected
    along the trajectory of the dynamical system of interest,
    the diversity of the resulting dataset
    (which may be characterised by the quantity of information)
    is conditioned on the subsequent states of the system.
    Informally, the setting in which the sampling procedure
    is constrained by the current system state may introduce
    information redundancy if the system exhibits
    high auto-correlation or if the current state is in a
    slowly evolving region of the state space.
    Indeed, as shown in Figure~\ref{fig:lorenz_autocorrelation_x3}
    (auto-correlation from a perturbated fixed point of a controlled Lorenz 63' system), the auto-correlation from an initial state can be high
    in average for a long period of time while the control intensity allows
    to reduce the correlation of the sequence of states.

    \begin{wrapfigure}[14]{r}{0.4\textwidth}
        \vspace{-10pt}
        \captionsetup{width=0.95\linewidth}
        \begin{center}
            \includegraphics[width=0.40\textwidth]
            {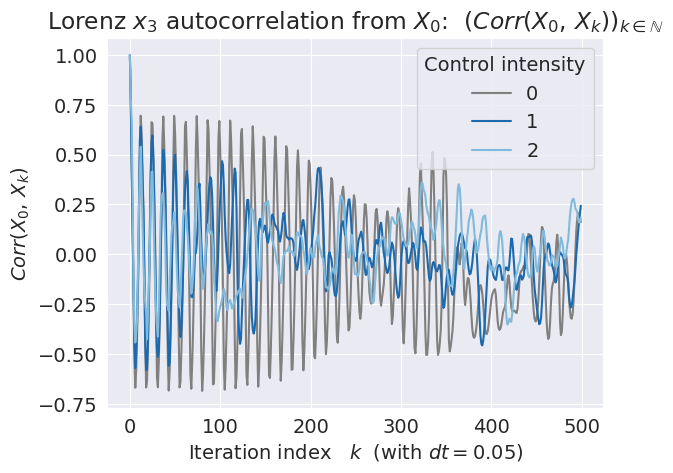}
        \end{center}
        \caption{
            $(Cov(\stateProcess_0,
            \stateProcess_\discreteTimeIndex))_{\discreteTimeIndex \in \mathbb{N}}$
            for the controlled Lorenz system $x_3$ component under multiple
            control intensities. }
        \label{fig:lorenz_autocorrelation_x3}
    \end{wrapfigure}

    However, for dynamical systems characterised by a broad range of
    time scales, the notion of \emph{temporal abstraction}, described in the below paragraphs,
    ~\citep{precup2000, machado2023} may play a key role in overcoming the issue mentioned here.

    \emph{Abstraction} in Artificial Intelligence refers to a broad range of
    techniques in order to provide a more compact representation
    of the problem at hand~\citep{boutilier1994,banse2023}.
    In the framework of \textit{Markov Decision Process (MDP)},
    the work of~\cite{sutton1999b} sheds light on the limitation
    induced by standard MDP modeling:
    \enquote{There is no notion of a course of action persisting over a variable
    period of time. \textelp{} As a consequence, conventional MDP methods are
    unable to take advantage of the simplicities and efficiencies sometimes
    available at higher levels of temporal abstraction.}

    \emph{Temporal abstraction} can refer to the concept of selecting the
    right level of time granularity to facilitate the description of the
    world model to achieve a given task.
    In simpler words, in the present case, temporal abstraction is the
    idea of representing and reasoning about actions and states at
    different time-scales and duration.

    In the present article, temporal abstraction through
    \textit{Semi-Markov Decision Processes (SMDP)} modeling is introduced
    to improve the informativeness of the sequential exploration of
    the state-control space.
    SMDP modeling is shown to obtain a better sample complexity
    of the dynamics model estimator.
    This article thus extends the previous work of~\cite{mehta2022b}
    by introducing temporal abstraction to the acquisition function.
    The paper is organised as follows.
    Section~\ref{sec:related-works} reviews the related works.  
    Section~\ref{sec:problem-setting} introduces the problem setting.
    Section~\ref{sec:method-experiments} presents the hypothesis and
    the experimental setup while Section~\ref{sec:results} presents the results
    and Section~\ref{sec:conclusion} concludes the paper.
    }


    \section[Related Works]{Related Works}\label{sec:related-works}

    \paragraph[Information Driven Model-Based Control]
    {Information Driven Model-Based Control}
    \label{par:information-driven-model-based-control}

    The foundations of the \textit{Bayesian Experimental Design} have been laid
    by the seminal work of \citet{lindley1956} where the author presents
    a measure of the information provided by an experiment.
    More recently, \citet{mackay1992} termed \textit{Expected Information Gain (EIG)} a measure of the information
    provided by an observation allowing, in his own terms,
    to \emph{actively select particularly salient data points}.
    In the field of LB-MPC, such a criterion has been used to cherry-pick
    the most informative state-control pair to learn the dynamics of
    the system~\citep{mehta2022, mehta2022b}.
    Their work is based on the broader Bayesian Optimisation method of
    ~\cite{neiswanger2021} designed to optimise ``blackbox'' functions.
    An extensive review of Bayesian Optimisation and its applications is
    available in this latter paper.

    \paragraph[Learning-Based Model Predictive Control]
    {Learning-Based Model Predictive Control}
    \label{par:learning-based-model-predictive-control}

    The history of learning-based models may be traced back to the seminal work
    by~\cite{stratonovich1960} in probability theory which stimulated
    several contributions, notably the work of~\cite{kalman1961},
    that were to compose a body of work generally referred to as filtering
    theory.
    More recently, ~\cite{kamthe2018} model the system dynamics
    with Gaussian Processes (GP) and use MPC for data
    efficiency.
    GPs are also used in the PILCO model ~\citep{deisenroth2011}
    which has a high influence in MB-RL\@.
    \cite{koller2019} model the uncertainty of the system dynamics for
    safe-RL\@.
    The work of~\cite{bonzanini2020} presents a stochastic LB-MPC
    strategy to handle this uncertainty.

    \paragraph[Semi-Markov Decision Processes]{Semi-Markov Decision Processes}
    \label{par:related-works-semi-markov-decision-processes}
    Temporal abstraction in reinforcement learning was pioneered
    in~\cite{sutton1995} and~\cite{precup1997, precup2000}.
    Specifically, ~\cite{sutton1995} proposed learning a model and value
    function at different levels of temporal abstraction.
    The actions in SMDPs take variable amounts of time and are intended to model
    temporally-extended courses of action.
    Recent works for continuous-time control use variants of
    Neural Ordinary Differential Equations to model dynamics delays
    ~\citep{du2020, holt2023}.
    A classical use of SMDP is for queueing control and
    equipment maintenance~\citep{puterman2014} where time-delays are
    prominent.

    \section[Problem Setting]{Problem Setting}\label{sec:problem-setting}
    \subsection[Control Model]{Control Model}\label{subsec:control-model}

    This work considers a control model given by the following $d$-dimensional
    discrete-time dynamical system $\stateProcess$~\citep{duflo1997} on
    a probability space
    $\left(
    \sampleSpace,
    \sampleSpaceSigmaAlgebra,
    \probabilityOperator
    \right)$
    defined by
    \begin{equation}
        \begin{aligned}
            & \stateProcess_{\discreteTimeIndex+1} =
            \dynamicsOperator
            \left(
            \stateProcess_{\discreteTimeIndex},\,
            \controlProcess_{\discreteTimeIndex},\,
            \disturbanceProcess_{\discreteTimeIndex}
            \right) \\
            & \stateProcess_0
            \sim \mathcal{N}(\startEquilibrium,\, \startEquilibriumStd^2 I_d)
        \end{aligned}
        \label{eq:dynamics-recurence-equation}
    \end{equation}
    with $\stateProcess_{\discreteTimeIndex} \in \stateSpace$,
    $\controlProcess_{\discreteTimeIndex} \in \controlSpace$
    and $\disturbanceProcess_{\discreteTimeIndex} \in \mathcal{Z}$
    for any $k \in \mathbb{N}$, where $\mathcal{X}$, $\controlSpace$
    and $\disturbanceSpace$ are respectively the corresponding state, control
    and disturbance spaces.
    The initial state starts from a reference state $\startEquilibrium$
    (a system equilibrium or \textit{fixed point}\footnote{
        In this work a fixed point is considered as a point
        of the state space $\startEquilibrium \in \stateSpace$ such that
        $\dynamicsOperator(\startEquilibrium, 0) = \startEquilibrium$.}) on which
    centered Gaussian noise with diagonal covariance is additively applied,
    $\stateProcess_0 \sim
    \mathcal{N}(\startEquilibrium,\, \startEquilibriumStd^2 I_d)$.
    The \textit{i.i.d.} random process
    $(\disturbanceProcess_\discreteTimeIndex)_{\discreteTimeIndex \in \mathbb{N}}$
    is such that $\disturbanceProcess_\discreteTimeIndex$ is independent of all
    previous states and controls for any $k \in \mathbb{N}$.
    The distribution of $\disturbanceProcess_\discreteTimeIndex$ for any
    $k \in \mathbb{N}$ is denoted by
    $\probabilityOperator_{\disturbanceProcess}$.
    Coupled with the dynamics, an instantaneous cost function
    $\costFunction:
    \stateProcess \times \controlProcess
    \rightarrow
    \mathbb{R}_+$
    is also given to define the control model.

    In the sequel, it will be convenient to define the control model as a
    \emph{Markov Control Model (MCM)}~\citep{hernandez-lerma1989}
    defined by the following
    transition probability $\stateTransitionProbability$ on
    $\stateSpace \times \controlSpace$:
    \begin{equation}
        \stateTransitionProbability(B_\stateSpace,\, (\statePoint, \controlPoint))
        = \int_{\disturbanceSpace} 1_{B_\stateSpace}
        \left(\dynamicsOperator(x, u, z)\right)
        \probabilityOperator_{\disturbanceProcess}(dz)
        = \probabilityOperator_{\disturbanceProcess}
        (\left\{ \disturbancePoint \in \disturbanceSpace
        \mid
        \dynamicsOperator(\statePoint, \controlPoint, \disturbancePoint)
        \in B_\stateSpace \right\})
        \label{eq:transition-kernel}
    \end{equation}
    for any $B_\stateSpace \in \mathcal{B}(\stateSpace)$
    (Borel $\sigma$-algebra) and
    $(\statePoint, \controlPoint)
    \in
    \stateSpace \times \controlSpace$.
    The function $1$ is the indicator function.

    Hence, the conditional distribution of $\stateProcess_{\discreteTimeIndex+1}$
    given $\stateProcess_\discreteTimeIndex$ and
    $\controlProcess_\discreteTimeIndex$ is given by
    \begin{equation}
        \stateTransitionProbability(B_\stateSpace, (\statePoint, \controlPoint))
        = \probabilityOperator(\stateProcess_{\discreteTimeIndex+1}
        \in
        B_\stateSpace \mid \stateProcess_\discreteTimeIndex = \statePoint,
        \controlProcess_\discreteTimeIndex = \controlPoint)
        \label{eq:conditional-distribution-state-transition-kernel}
    \end{equation}

    Additionally, in this context, a \emph{policy} $\policy$ is a transition
    probability on $\mathcal{U}$ given $\mathcal{X}$, \textit{i.e.},
    a distribution on controls conditioned on states.
    In the rest of the paper,
        $\policy\left( \statePoint, d\controlPoint \right) = \delta_{\left\{ \controlPoint \right\}}$
        is the Dirac measure at $\controlPoint$.
        Hence the notation is simplified to $\policy\left( \statePoint \right) = \controlPoint$.

    Together, a control model, a policy $\policy$ and an initial distribution on
    $\stateSpace$ define a stochastic process with distribution
    $\probabilityOperator^{\policy}$ on the space of trajectories
    $(\stateSpace \times \controlSpace)^{\timeHorizon}$.
    The distribution of the process is given by
    $\probabilityOperator(d{\statePoint_0}d{\controlPoint_0}d{\statePoint_1}
    \dots )
    = \probabilityOperator_{X_0}(d{\statePoint_0})
    \policy(d{\controlPoint_0}
    \mid d{\statePoint_0})
    \stateTransitionProbability(d{\statePoint_1}
    \mid d{\statePoint_0}, d{\controlPoint_0})
    \dots
    $
    More details on the stochastic process are given
    in~\cite{hernandez-lerma1996, puterman2014}.
    Lastly, the history process
    $(\historyProcess_{\discreteTimeIndex})_{\discreteTimeIndex \in \mathbb{N}}$
    is defined as
    $\historyProcess_{\discreteTimeIndex}
    = (\stateProcess_0,\, \controlProcess_0, \dots,
    \stateProcess_\discreteTimeIndex)$
    for any $\discreteTimeIndex \in \mathbb{N}$.
    When $\discreteTimeIndex = \timeHorizon$, $\historyProcess_{\timeHorizon}$
    is called the \emph{trajectory} of the process.
    The process $(\stateProcess_\discreteTimeIndex,
    \controlProcess_\discreteTimeIndex,
    \stateProcess_{\discreteTimeIndex+1})_{\discreteTimeIndex \in \mathbb{N}}$
    is called the \emph{transition process} and the marginal process
    $\left(\stateProcess_\discreteTimeIndex \right)_{\discreteTimeIndex \in \mathbb{N}}$
    is called a \emph{Markov Decision Process (MDP)}.

    \subsection[Control Problem]{Control Problem}
    \label{subsec:control-problem}

    The studied control problem is to find a policy $\policy^*$ which minimises
    the following performance criterion
    \begin{equation}
        \objectiveFunciton^{\policy} = \mathbb{E}^{\policy}
        \left[\sum_{\discreteTimeIndex=0}^{\timeHorizon}
        \costFunction\left(\stateProcess_{\discreteTimeIndex},
        \controlProcess_{\discreteTimeIndex}\right)
        \right]
        \label{eq:finite-time-horizon-policy-gradient-objective}
    \end{equation}
    where $\timeHorizon \in \mathbb{N}$ is a given time-horizon and
    $\mathbb{E}^{\policy}$ denotes the expectation under the probability measure
    $P^{\policy}$.
    The quantity $\objectiveFunciton^{\policy}$ is called the value function or
    objective function.
    The history process under $\policy^*$ is called the optimal history process
    and is denoted by $(\historyProcess_{\discreteTimeIndex}^*)_{\discreteTimeIndex
    \in \mathbb{N}}$ and the random variable $\historyProcess_{\timeHorizon}^*$
    is called the \emph{optimal trajectory}.

    In this work, the optimal policy $\policy^*$ is estimated with
    \textit{Model Predictive Control (MPC)} applied on a model of the dynamics.
    The MPC approach consists of solving a
    finite-horizon optimal control problem at each time step, formally it
    defines the following policy
    \begin{flalign}
        & \MPCpolicy(\statePoint) = \controlPoint_0^* \\
        & s.t. \quad
        (\controlPoint_0^*,\, \ldots,\, \controlPoint_{\timehorizonMPC}^*) =
        \argmin_{
            (\controlPoint_0,\,
            \ldots,\,
            \controlPoint_{\timehorizonMPC})}
        \mathbb{E}^{
            (\controlPoint_0,\,
            \ldots,\,
            \controlPoint_{\timehorizonMPC})}
        \left[ \sum_{\discreteTimeIndex=0}^{\timehorizonMPC}
        \costFunction
        \left(
        \stateProcess_{\discreteTimeIndex},
        \controlPoint_{\discreteTimeIndex}
        \right)
        \mid
        \stateProcess_0 = \statePoint
        \right]
        \label{eq:model-predictive-control-policy}
    \end{flalign}
    where $\timehorizonMPC \leq \timeHorizon$ is the MPC planning horizon,
    $\statePoint \in \stateSpace$ is the current state and the expectation is taken with the underlying probability measure
    $P^{(\controlPoint_0,\, \ldots,\, \controlPoint_{\timehorizonMPC})}$ is
    characterised by a deterministic policy (Dirac measures)
    concentrated at $u_\discreteTimeIndex$ for all
    $0 \leq k \leq \timehorizonMPC$.
    The policy obtained with MPC on $\stateTransitionProbability$
    is denoted by $\MPCpolicy$.
    The history process under $\MPCpolicy$ is denoted by
    $\historyProcess^{\text{MPC}} =
    (\historyProcess_{\discreteTimeIndex}^{\text{MPC}})
    _{\discreteTimeIndex \in \mathbb{N}}$, it is an
    approximation of the optimal history process
    $(\historyProcess_{\discreteTimeIndex}^*)_{\discreteTimeIndex\in \mathbb{N}}$
    and the random variable
    $\historyProcess_{\timeHorizon}^\text{MPC}$ is an approximation of the
    optimal trajectory $\historyProcess_{\timeHorizon}^*$.
    The objective function under $\MPCpolicy$ is denoted by
    $\objectiveFunciton^{\text{MPC}}$.
    The MPC procedure is here performed with the \textit{iCEM} algorithm,
    an improved version of the \textit{Cross Entropy Method (CEM)}
    ~\citep{rubinstein2004, pinneri2021}, a zeroth-order optimisation algorithm
    based on Monte-Carlo estimation.

    \subsection[Gaussian Process Modeling]{Gaussian Process Modeling}
    \label{subsec:gaussian-process-modeling}
    The use of \textit{Gaussian Process (GP)} regression to model relevant
    quantities of controlled dynamical systems has long been proposed~\citep{kuss2003, deisenroth2011, kamthe2018} notably for its
    distributional nature thus its ability to model uncertainty.
    By definition, a GP is a stochastic process (here indexed by
    $\stateSpace \times \controlSpace$) such that any finite collection of
    random variables has a joint Gaussian distribution.

    Continuing from the aforementioned papers, GP regression is used to model
    the transition probability $\stateTransitionProbability$ with a
    model estimator $\hat{\stateTransitionProbability}_{\stateControlDataset}$
    such that
    \begin{equation}
        \hat{\stateTransitionProbability}_{\stateControlDataset}
        (~ \cdot ~,\, (\statePoint, \controlPoint))
        \sim
        \mathcal{N}
        \left(
        \GPmean(\statePoint, \controlPoint),\,
        \GPcovariance
        \left(
        (\statePoint, \controlPoint),\,
        (\statePoint, \controlPoint)
        \right)
        \mid
        \stateControlDataset\right)
        \label{eq:conditional-distribution-gaussian-process}
    \end{equation}
    where $\GPmean$ and $\GPcovariance$ are respectively the mean and covariance
    functions of the GP\@ and
    $\stateControlDataset$ is a dataset of observations from
    the transition process
    $(\stateProcess_\discreteTimeIndex, \controlProcess_\discreteTimeIndex,
    \stateProcess_{\discreteTimeIndex+1})_{\discreteTimeIndex \in \mathbb{N}}$.
    The distribution $\hat{\stateTransitionProbability}_{\stateControlDataset}$ of Equation
    ~\eqref{eq:conditional-distribution-gaussian-process} is the predictive
    posterior distribution of the GP conditioned on the dataset
    $\stateControlDataset$ (the reader is referred to~\cite{rasmussen2006}
    for more details on GP regression).
    The processes $\hat{\stateProcess}$, $\hat{\controlProcess}$ and
    $\hat{\historyProcess}$ are respectively the state, control and history
    processes under the approximate model and the same rules of notation apply
    as for the original processes.
    The MPC policy obtained with the approximate model
    $\hat{\stateTransitionProbability}_{\stateControlDataset}$ is denoted by
    $\MPCpolicyEstimator$.
    The history process under $\MPCpolicyEstimator$ is
    denoted by $\widehat{\historyProcess}^{\text{MPC}} =
    (\widehat{\historyProcess}_{\discreteTimeIndex}^{\text{MPC}})
    _{\discreteTimeIndex \in \mathbb{N}}$ and the objective function under
    $\MPCpolicyEstimator$ is denoted by $\widehat{\objectiveFunciton}^{\text{MPC}}$.


    Notably, \emph{this work focuses on the sample complexity required to
    estimate a model $\hat{\stateTransitionProbability}_{\stateControlDataset}$
        of the true dynamics $\stateTransitionProbability$ accurate enough
        to obtain a MPC policy $\MPCpolicyEstimator$ that is close to the
        optimal policy $\policy^*$.}

    Hence, two time units are considered: the sampling iteration
    $\samplingIteration$ which represents the number of observations gathered
    from the system so far, and the time index $\discreteTimeIndex$ of the current
    state $\stateProcess_\discreteTimeIndex$
    of the underlying dynamical system $\stateProcess$\@.
    It is supposed in the following that
    $\samplingIteration \leq \discreteTimeIndex$: it is not possible to gather
    more observations than the number of time steps of the system.


    \subsection[Expected Information Gain]{Expected Information Gain}
    \label{subsec:expected-information-gain}
    For a fixed sampling budget $\samplingIteration$ and a fixed configuration
    (\textit{e.g.} the horizon $\timehorizonMPC$, the number of samples for
    the Monte-Carlo estimation of the cost or the other hyper-parameters of the
    iCEM algorithm)
    to perform the MPC procedure $\MPCpolicy$,
    the control performance mainly lies in the quality of the model estimator
    $\hat{\stateTransitionProbability}_
        {\stateControlDataset_\samplingIteration}$.
    It depends on two main elements: the choice a priori of the mean and
    kernel functions $\mu$ and $\GPcovariance$ and the collection
    $\stateControlDataset_\samplingIteration$ of $\samplingIteration$
    observations.
    From the work of \citet{mehta2022, mehta2022b}, the selection of the
    observations can be guided by the maximisation of the
    \textit{Expected Information Gain (EIG)} on the
    optimal trajectory. 

    Let suppose the time iteration $\discreteTimeIndex$ of the underlying
    observed process $\stateProcess$ is equal to the number of samples gathered,
    \textit{i.e.}, $\discreteTimeIndex = \samplingIteration$ and the dataset is
    already collected\footnote{In this specific case of
        $\discreteTimeIndex = \samplingIteration$, the dataset
        $\stateControlDataset_\samplingIteration$ simply contains the whole past
        trajectory of $\stateProcess$, it is a realisation of
        $\historyProcess_\samplingIteration$, in other words
        $\stateControlDataset_\samplingIteration
        = \historyProcess_\samplingIteration(\omega)$ for some random outcome
        $\omega \in \Omega$.}
    at the sampling iteration $\samplingIteration$ such that
    $\stateControlDataset_\samplingIteration =
    \left((\statePoint_i, \controlPoint_i, {\statePoint '}_i)\right)
    _{i=0}^{\samplingIteration - 1}$
    and denote by
    $\left(
    \stateProcess_{\samplingIteration},
    \controlProcess_{\samplingIteration}
    \right)$
    a new random state-control pair to draw from the system.
    The goal is to select the state-control pair
    $(\statePoint, \controlPoint)$ that maximises the \textit{Expected Information Gain} $\expectedInformationGain$ on the optimal trajectory which is defined by
    \begin{equation}
        \expectedInformationGain_\samplingIteration
        (\statePoint, \controlPoint) =
        \entropy
        \left[
            \hat{\historyProcess}_\timeHorizon^*
            \mid  \stateControlDataset_\samplingIteration \right]
        -
        \mathbb{E}_{\probabilityOperator_{\stateProcess_{n+1}
        \mid
        \stateControlDataset_\samplingIteration,\,
        \stateProcess_\samplingIteration = \statePoint,\,
        \controlProcess_\samplingIteration = \controlPoint}}
        \left[
            \entropy
            \left[
                \hat{\historyProcess}_\timeHorizon^*
                \mid \stateControlDataset_\samplingIteration,
                \stateProcess_\samplingIteration = \statePoint,
                \controlProcess_\samplingIteration = \controlPoint,
                \stateProcess_{n+1}
                \right]
            \right]
        \label{eq:expected-information-gain-trajectory-definition}
    \end{equation}
    where $\entropy$ denotes the differential entropy of a random variable.
    In other words, given a level of uncertainty
    $\entropy\left[\hat{\historyProcess}_\timeHorizon^*
    \mid  \mathcal{D}_\samplingIteration\right]$
    on the optimal trajectory $\hat{\historyProcess}_\timeHorizon^*$,
    the EIG measures the reduction of this uncertainty when the dataset of the
    model estimator is augmented with the transition tuple
    ($\statePoint, \controlPoint, \stateProcess_{n+1}$).

    An intriguing interpretation can be made by noticing that~\eqref{eq:expected-information-gain-trajectory-definition} is also equal to the negative \emph{Conditional Mutual Information (CMI)}~\citep{pinsker1964, cover2006} of the optimal trajectory $\hat{\historyProcess}_\timeHorizon^*$ and the new state $\stateProcess_{\samplingIteration+1}$ given the dataset $\stateControlDataset_\samplingIteration$ and the state-control pair $(\stateProcess_{\samplingIteration}, \controlProcess_{\samplingIteration})$~\footnote{Here and after, a slight abuse of notation is made as the dataset $\stateControlDataset_\samplingIteration$ should be written $\stateControlDataset_\samplingIteration = \left((\statePoint_i, \controlPoint_i, {\statePoint '}_i)\right)_{i=0}^{\samplingIteration - 1}$ since the sole random quantities are $\stateProcess_{\samplingIteration}$ and $\hat{\historyProcess}_\timeHorizon^*$ but it is omitted for the sake of readability.}.
    Thus, maximising the EIG is equivalent to minimising the
    CMI between the
    optimal trajectory and the new transition tuple hence tending to draw
    new states sharing less information with the optimal trajectory conditioned
    on the dataset $\mathcal{D}_\samplingIteration$ and the event
    ($\stateProcess_\samplingIteration = x,
    \controlProcess_\samplingIteration = u$).
    Indeed, by definition, the CMI quantifies the
    independence between the distribution of the optimal trajectory and the
    distribution of the new state given both the dataset and the current state-action
    pair.
    \\
    \indent By symmetry of the EIG, a more tractable formulation is given by
    \vspace{-0.2cm}
    \begin{equation}
        \expectedInformationGain_\samplingIteration
        (\statePoint, \controlPoint) =
        \entropy
        \left[
            \stateProcess_{n+1}
            \mid
            \stateControlDataset_\samplingIteration,\,
            \stateProcess_\samplingIteration = \statePoint,\,
            \controlProcess_\samplingIteration = \controlPoint
            \right]
        - \mathop{\mathbb{E}}_{\probabilityOperator_{\hat{\historyProcess}_T^*
        \mid
        \stateControlDataset_\samplingIteration}}
        \left[
            \entropy
            \left[
                \stateProcess_{n+1}
                \mid
                \stateControlDataset_\samplingIteration,
                \stateProcess_\samplingIteration = \statePoint,\,
                \controlProcess_\samplingIteration = \controlPoint,\,
                \hat{\historyProcess}_T^*
                \right]
            \right]
        \label{eq:expected-information-gain-nextstate-definition}
    \end{equation}
    It is in practice estimated by Monte-Carlo sampling as detailed in
    Section~\ref{sec:method-experiments}.

    In the original work of \citet{mehta2022}, the EIG is maximised with a
    greedy Monte-Carlo algorithm (uniform sampling) that selects the next
    state-control pair $(\statePoint, \controlPoint)$ to interact with the true
    system and subsequently update the dataset
    $\stateControlDataset_\samplingIteration$ with the
    new transition tuple $(\statePoint,\, \controlPoint,\, \statePoint')$ where
    $\statePoint'$ is sampled from the true transition probability
    $\stateTransitionProbability(~ \cdot ~, (\statePoint, \controlPoint))$.
    It assumes any state-control pair $(\statePoint, \controlPoint)$ can be
    evaluated and queried at any time step.
    The authors' algorithm is called \textit{Bayesian Active
    Reinforcement Learning (BARL)}; the dataset and EIG obtained with this
    algorithm are denoted by
    $\stateControlDataset_\samplingIteration^{\text{BARL}}$ and
    $\expectedInformationGain^{\text{BARL}}$ respectively.
    In this setting, the dataset support is the whole state-control space,
    $\support(\stateControlDataset_\samplingIteration^{\text{BARL}})
    = (\stateSpace \times \controlSpace \times \stateSpace)^n$.

    However, in many real-world applications, the system is not always
    controllable and the state-control pairs that can
    be queried are limited to a subset induced by the system trajectory.
    This constraint has been considered in the work following the original
    paper~\citep{mehta2022b} where
    the authors proposed to restrict the dataset
    support to the trajectory of the system.
    This second algorithm is called\footnote{It is important to mention that the main asset of TIP is to provide a whole trajectory as input to the EIG, which is not used in this work.
    Thus, only the property of querying observation by following the trajectory of the system is used here.} \textit{Trajectory Information Planning (TIP)} and similarly the dataset and EIG obtained with this algorithm are denoted by
    $\stateControlDataset_\samplingIteration^{\text{TIP}}$
    and
    $\expectedInformationGain^{\text{TIP}}$ respectively.

    In this case, the dataset support is limited to the trajectory of
    the system,
    $\support(\stateControlDataset_\samplingIteration^{\text{TIP}})
    \subseteq
    \{((\statePoint_k,\, \controlPoint_k,\, \statePoint_{k+1}))
    _{k=1}^{\samplingIteration}
    \in
    (\stateSpace \times \controlSpace \times \stateSpace)^n
    \mid
    \exists (\disturbancePoint_k)_{k=1}^n
    \in \disturbanceSpace^n,\,
    \statePoint_{k+1} =
    \dynamicsOperator
    (\statePoint_k,\, \controlPoint_k,\, \disturbancePoint_k),\,
    0 \leq i \leq n \}
    \subseteq
    (\stateSpace \times \controlSpace \times \stateSpace)^n
    = \support(\stateControlDataset_\samplingIteration^{\text{BARL}})$.
    This set inclusion implies that the optimal EIG obtained with TIP is lower
    than the one obtained with BARL provided the transition probability
    estimator $\hat{\stateTransitionProbability}
    _{\stateControlDataset_\samplingIteration}$ are
    the same for both algorithms for a fixed current state
    $\statePoint \in \stateSpace$,
    $\max_{\{(\statePoint, \controlPoint),\,
    \controlPoint' \in \controlSpace\}}
    \expectedInformationGain(\statePoint, \controlPoint')
    \leq
    \max_{\{(\statePoint', \controlPoint')
    \in \stateSpace \times \controlSpace\}}
    \expectedInformationGain(\statePoint', \controlPoint')$.

    Besides, the latter algorithm (TIP) do not take into account the potential benefits of including dynamics time scales in the sampling process.
    In the next section, an extension of the TIP algorithm is proposed
    to increase the EIG for each of the sampling iteration through the
    \emph{introduction of delayed state-control pairs in the setting of}
    \textit{Semi-Markov Decision Processes (SMDP)}.
    The new algorithm builds upon TIP by considering the inclusion of temporally-extended actions in the data-collection procedure to reach more distant system states that are not reachable with the original TIP algorithm, hence increasing the amount of information gathered from the system.
    A similar use of action repetition improves learning in Deep-RL~\citep{sharma2017,lakshminarayanan2017}.

    \subsection[Semi-Markov Decision Processes Extension]{Semi-Markov Decision Processes Extension}
    \label{subsec:semi-markov-decision-processes-extension}

%
    A formal definition of \emph{temporal abstraction} is given through the concept of \emph{options} defined by~\cite{sutton1999b} where it refers to \emph{temporally extended courses of action}.
    This concept has been shown by~\cite{parr1998} to be equivalent to the construction of \emph{Semi-Markov Decision Processes (SMDP)} which are defined below.

    Let call \emph{decision epoch} the time index $\discreteTimeIndex$ of the
    underlying dynamics
    $(\stateProcess_\discreteTimeIndex)_{\discreteTimeIndex \in \mathbb{N}}$
    defined by equation~\eqref{eq:dynamics-recurence-equation}.
    \emph{Semi-Markov Control Models (SMCM)} generalise the concept of MCM by
    letting the decisions be random variables.
    Indeed, consider a strictly increasing random sequence
    $(\randomDiscreteTime_j)_{j \in \mathbb{N}}$
    of integers.
    The random quantities $\randomInterdecisionTime_j
    = \randomDiscreteTime_j - \randomDiscreteTime_{j-1}$
    with support in some finite space $\timeDelaySpace \subsetneq \mathbb{N} \setminus \left\{ 0 \right\}$ are called
    \emph{inter-decision times} and the random index $\randomDiscreteTime_j$ are
    called \emph{random decision epochs}.
    The resulting stochastic process
    $(\stateProcess_{\randomDiscreteTime_j})_{j \in \mathbb{N}}$ is called a
    \emph{semi-Markov Decision Process}.
    For a more detailed probabilistic construction,
    see~\cite[p.~534]{puterman2014} and~\cite[p.~15]{hernandez-lerma1989}.

    In the scope of this paper, \emph{SMDP are used to model the temporal extension of
    the control process}.
    The corresponding SMCM is introduced by first extending the control
    space from $\controlSpace$ to $\controlSpace \times \timeDelaySpace$ such
    that the temporal extension of the control is encoded in the last coordinate
    of the control tuple,
    and the new dynamics is given by
    $\stateTransitionProbability^{\text{SMDP}}(d\statePoint'
    \mid
    (\statePoint, (\controlPoint, \timeDelayPoint)))
    = \probabilityOperator(
    \stateProcess_{\discreteTimeIndex + \timeDelayPoint}
    \mid
    \stateProcess_\discreteTimeIndex = \statePoint,\,
    \controlProcess_{\discreteTimeIndex:\discreteTimeIndex + \timeDelayPoint - 1}
    = \controlPoint)$ where
    $\controlProcess_{\discreteTimeIndex:\discreteTimeIndex + \timeDelayPoint - 1}
    = \controlPoint$ means that \emph{the control process is constant} between
    $\discreteTimeIndex$ and $\discreteTimeIndex + \timeDelayPoint - 1$.
    The latter definition illustrates the fact that during the inter-decision
    time $\randomInterdecisionTime = \timeDelayPoint$, the control process is
    constant and equal to $\controlPoint$.

    From now on, this construction allows to enlarge the support of the dataset
    $\stateControlDataset_\samplingIteration$, for a fixed number of
    observations $n$ while maintaining a rollout, trajectory-based sampling
    procedure.
    Indeed, the dataset support is now
    $\support(\stateControlDataset_\samplingIteration^\SMTIP) \subseteq
    \{
    ((\statePoint_{\discreteTimeIndex_j},\,
    \controlPoint_{\discreteTimeIndex_j},\,
    \statePoint_{\discreteTimeIndex_j+1}))_{{j}=1}^n
    \in (\stateSpace \times \controlSpace \times \stateSpace)^n
    \mid
    \exists
    (\disturbancePoint_k)_{k=1}^{\samplingIteration \sup(\timeDelaySpace)}
    \in \disturbanceSpace^{\samplingIteration \sup(\timeDelaySpace)},\,
    \statePoint_{k + 1} = \dynamicsOperator
    (\statePoint_k, \controlPoint_k, \disturbancePoint_k),\,
    0 \leq k \leq \samplingIteration \sup(\timeDelaySpace),\,
    (\discreteTimeIndex_j)_{j=1}^{\samplingIteration}
    \in
    \timeDelaySpace^{\samplingIteration},\,
    \discreteTimeIndex_j < \discreteTimeIndex_{j+1}
    \}$, the transitions tuples extracted from the set of all possible
    subsequences of the trajectory up to the maximal reachable time value.

    Therefore, $\support(\stateControlDataset_\samplingIteration^\TIP)
    \subseteq
    \support(\stateControlDataset_\samplingIteration^\SMTIP)$.
    Consequently, this suggests an extension of the EIG to the SMDP setting.
    Let $\timeDelayPoint \in \timeDelaySpace$ be an
    inter-decision time and $\stateControlDataset_\samplingIteration^\SMTIP$
    be the dataset under the SMDP setting at the sampling iteration $\samplingIteration$,
    the resulting $\expectedInformationGain_\samplingIteration^\SMTIP
    (\statePoint, (\controlPoint, \timeDelayPoint))$  is defined as
    \small
    \begin{equation}
        \entropy
        \!
        \left[
            \stateProcess_{
                \randomDiscreteTime_\samplingIteration \! + \timeDelayPoint + 1}
            \! \!
            \mid
            \!
            \stateControlDataset_{\samplingIteration},\,
            \stateProcess_{\randomDiscreteTime_\samplingIteration}
            \!
            = \statePoint,
            \controlProcess_{\randomDiscreteTime_\samplingIteration: \randomDiscreteTime_\samplingIteration + \timeDelayPoint}
            \!
            = \controlPoint,
            \randomDiscreteTime_{\samplingIteration}
            \right]
        - \mathbb{E}_{\probabilityOperator_{\hat{\historyProcess}_T^*
        \mid
        \stateControlDataset_{\samplingIteration}}} \hspace{-5pt}
        \left[
            \entropy \hspace{-2pt}
            \left[
                \!
                \stateProcess_{
                    \randomDiscreteTime_{\samplingIteration} +\, \timeDelayPoint + 1}
                \!
                \mid
                \!
                \stateControlDataset_{\samplingIteration},
                \stateProcess_{\randomDiscreteTime_\samplingIteration}
                \! \!
                = \statePoint,
                \controlProcess_{\randomDiscreteTime_\samplingIteration:\randomDiscreteTime_\samplingIteration +\, \timeDelayPoint}
                \!
                = \controlPoint,
                \hat{\historyProcess}_T^*,
                \randomDiscreteTime_{\samplingIteration}
                \!
                \right]
            \!
            \right]
        \label{eq:expected-information-gain-nextstate-definition-semimarkov}
    \end{equation}
    \normalsize
    Hence, this measure allows the introduction of temporal abstraction in the
    sampling procedure by considering the inter-decision delay to increase the
    potential information gain.
    However, despite being tractable in trajectory rollout settings, the
    metric defined by~\eqref{eq:expected-information-gain-nextstate-definition-semimarkov}
    needs to look ahead in the future to be computed (non-causal).
    Last, note that
    $\expectedInformationGain^{\SMTIP}(\statePoint, \controlPoint, 1)
    = \expectedInformationGain^{\TIP}(\statePoint, \controlPoint)$.

    \section[Method and Experiments]{Method and Experiments}
    \label{sec:method-experiments}
    The main objective of this work is to demonstrate the increase in the total
    information gathered from a system with the introduction of temporal
    abstraction via the $\expectedInformationGain^\SMTIP$ measure.
    To this end, a comparison between the original TIP algorithm and the
    proposed SMDP extension is performed on two controlled dynamical systems,
    the Inverted Pendulum~\citep{trelat2005} and
    the Lorenz Attractor~\citep{vincent1991}.

    The algorithm controls the path of the dynamical system
    $\left( \stateProcess_\discreteTimeIndex
    \right)_{\discreteTimeIndex \in \mathbb{N}}$ and collects observations
    $\left( \stateProcess_i,\, \controlProcess_i,\, \stateProcess_{i+1} \right)
    _{i=0}^{\samplingIteration - 1}$
    to populate the dataset $\stateControlDataset_\samplingIteration$ and
    improve the GP transition probability estimator
    $\hat{\stateTransitionProbability}_{\stateControlDataset_\samplingIteration}$.
    The indices $\samplingIteration$ and $\discreteTimeIndex$ are respectively the
    sampling iteration and the time index of the underlying dynamical system
    $\left( \stateProcess_\discreteTimeIndex \right)
    _{\discreteTimeIndex \in \mathbb{N}}$.
    The TIP algorithm supposes $\samplingIteration =
    \discreteTimeIndex$ (data collected at each time step) while
    $\samplingIteration \leq \discreteTimeIndex$ (there are time steps where
    no data is collected) for the SMDP extension.
    In the SMDP case, the inter-decision time
    $\randomInterdecisionTime_\samplingIteration$ rules the optional sampling
    procedure which defines the random decision epochs
    $\randomDiscreteTime_n = \randomDiscreteTime_{\samplingIteration-1} +
    \randomInterdecisionTime_\samplingIteration$.
    The random decision epochs $\randomDiscreteTime_\samplingIteration$ define when the
    algorithm can query the system
    $\left( \stateProcess_\discreteTimeIndex \right)
    _{\discreteTimeIndex \in \mathbb{N}}$.

    To estimate $\expectedInformationGain_\samplingIteration^{\SMTIP }$, a collection of
    bootstrapped future states, candidate control points and inter-decision
    times are sampled.
    The bootstrapped future states
    $\stateProcess_{\randomDiscreteTime_\samplingIteration + \timeDelayPoint}
    = \statePoint_\timeDelayPoint$ are estimated with the GP model.
    This may lead to a bias in the estimation of the EIG due to the
    bootstrapping error.
    The candidate control points and inter-decision times
    $(\controlPoint,\, \timeDelayPoint)$ are sampled from a uniform
    distribution $U(\controlSpace \times \timeDelaySpace)$
    at time $\randomDiscreteTime_\samplingIteration$
    to solve
    $\argmax_{
        (\controlPoint,\, \timeDelayPoint)
        \in \controlSpace \times \timeDelaySpace}
    \expectedInformationGain_\samplingIteration^{\SMTIP }
    (\statePoint_\timeDelayPoint, (\controlPoint, \timeDelayPoint))$.
    In this work, $\timeDelaySpace = \{1, \ldots, \timeDelayPoint_{\max}\}$
    for some $\timeDelayPoint_{\max} \in \mathbb{N}$.
    The $\expectedInformationGain_\samplingIteration^{\SMTIP}$ is estimated
    by the Monte-Carlo estimator
    $\widehat{\expectedInformationGain}_\samplingIteration^{\SMTIP}(\statePoint,
    (\controlPoint, \timeDelayPoint))$ given by
    \small
    \begingroup
    \setlength\abovedisplayskip{1pt}
    \begin{equation}
        \entropy
        \left[
            \stateProcess_{
                \randomDiscreteTime_\samplingIteration + \timeDelayPoint + 1}
            \mid
            \stateControlDataset_{\samplingIteration},
            \stateProcess_{\randomDiscreteTime_\samplingIteration}
            = \statePoint,
            \controlProcess_{\randomDiscreteTime_\samplingIteration + \timeDelayPoint}
            = \controlPoint,
            \randomDiscreteTime_{\samplingIteration}
            \right]
        - \frac{1}{m} \sum_{i=1}^m
        \entropy \hspace{-2pt}
        \left[
            \stateProcess_{
                \randomDiscreteTime_{\samplingIteration} +\, \timeDelayPoint + 1}
            \mid
            \stateControlDataset_{\samplingIteration},
            \stateProcess_{\randomDiscreteTime_\samplingIteration}
            = \statePoint,
            \controlProcess_{\randomDiscreteTime_\samplingIteration +\, \timeDelayPoint}
            = \controlPoint,
            {\hat{\historyProcess}_{T_i}^\text{MPC}},
            \randomDiscreteTime_{\samplingIteration}
            \right]
        \label{eq:expected-information-gain-nextstate-estimator-semimarkov}
    \end{equation}
    \endgroup
    \normalsize
    where $m$ is the number of Monte-Carlo samples of the optimal trajectory
    $\hat{\historyProcess}_{T_i}^\text{MPC}$ under $\hat{\stateTransitionProbability}
    _{\stateControlDataset_\samplingIteration}$.
    The entropy values are easily computed since the conditional distribution of the new state given the dataset and the current state-control pair is a Gaussian distribution with mean and covariance given by the GP posterior.
    More details on this procedure and the settings used are available in the
    paper of~\cite{mehta2022b}.

    Every two sampling iterations $\samplingIteration$, the MPC policy
    $\MPCpolicyEstimator$ is evaluated on the true system and the objective function
    is computed.
    Four independent experiments with different maximal inter-decision time
    $\timeDelayPoint_{\max} \in \left\{ 1, 2, 4, 8 \right\}$ are performed.
    For each of the experiments, the algorithm is run for 10 independent trials
    (seeds) to alleviate the variability proper to data-driven control
    methods~\citep{henderson2018}.
    The cost function is defined as
    $\costFunction(\statePoint, \controlPoint) = \|x \|^2$ in the case of the
    Lorenz attractor while the classic Gym~\citep{brockman2016} cost function
    (also norm-based) is used for the Inverted Pendulum.
    The sampling budget is set to $\samplingIteration_{\max} = 100$ for the
    Lorenz system and $\samplingIteration_{\max} = 200$ for the
    Inverted Pendulum.
    To implement the SMDP, the system is stepped forward in time with the action kept constant during inter-decision times.
    Details on the implementation and experimental settings are available on
        {\small\url{https://github.com/ReHoss/lbmpc_semimarkov}}.


    \section[Results]{Results}\label{sec:results}

    Among the relevant quantities to be reported,
    the evolution of the EIG, the interdecision times and the evaluation of the objective function
    are of interest to question the hypothesis raised in
    Section~\ref{sec:method-experiments}.

\begin{figure}[ht]
        \centering     
        \subfigure[Lorenz]
        {\label{fig:eig_evolution_lorenz}
        \includegraphics[width=60mm]{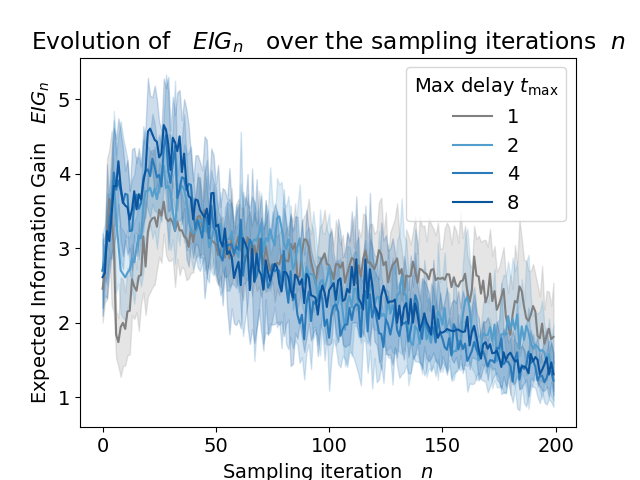}}
        \subfigure[Pendulum]
        {\label{fig:eig_evolution_pendulum}
        \includegraphics[width=60mm]{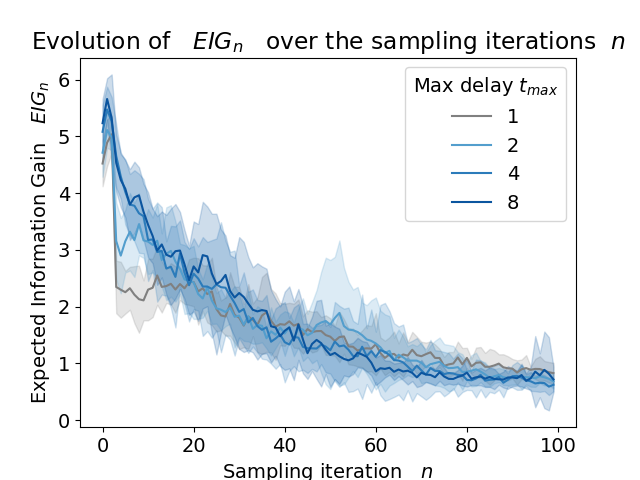}}
        \caption{Evolution of the Expected Information Gain
            $\expectedInformationGain^\SMTIP$ over the number of sampling iterations.}
        \label{fig:eig_evolution}
    \end{figure}

\begin{figure}[ht]
        \centering
        \subfigure[Lorenz]
        {\label{fig:interdecision_time_evolution_lorenz}
        \includegraphics[width=60mm]
        {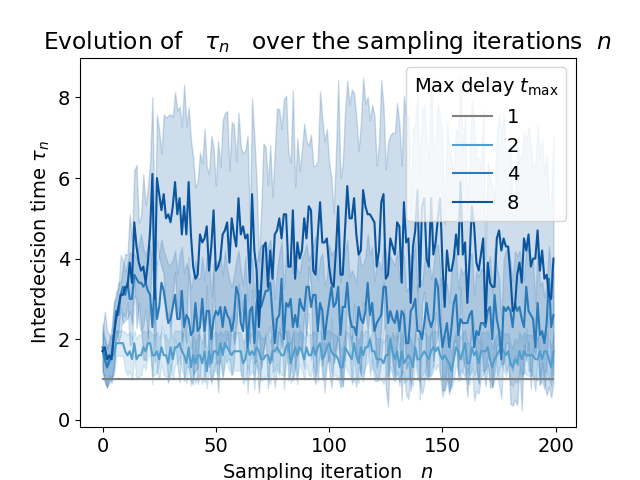}}
        \subfigure[Pendulum]
        {\label{fig:interdecision_time_evolution_pendulum}
        \includegraphics[width=60mm]
        {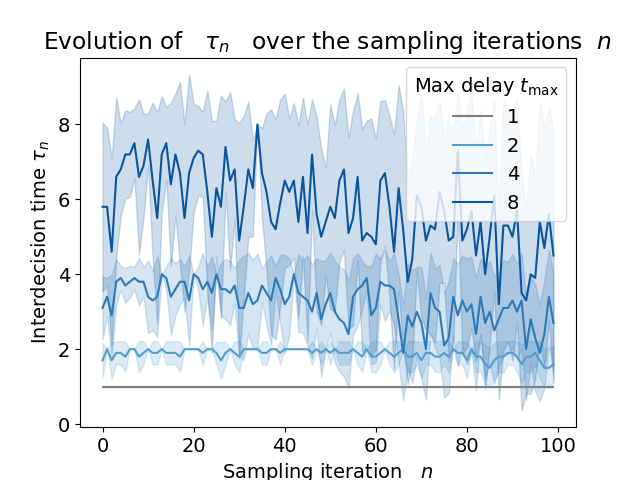}}
        \caption{Inter-decision time $\randomInterdecisionTime$ chosen by
        the SMDP during training.}
        \label{fig:interdecision_time_evolution}
    \end{figure}

\begin{figure}
        \centering
        \subfigure[Lorenz]
        {\label{fig:objective_function_evolution_lorenz}
        \includegraphics[width=60mm]
        {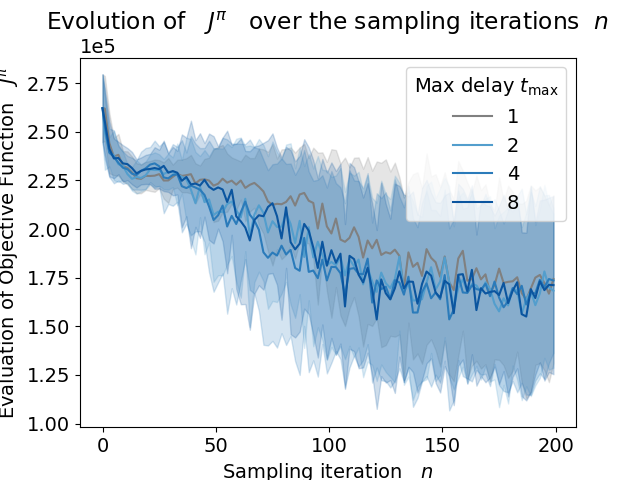}}
        \subfigure[Pendulum]
        {\label{fig:objective_function_evolution_pendulum}
        \includegraphics[width=60mm]
        {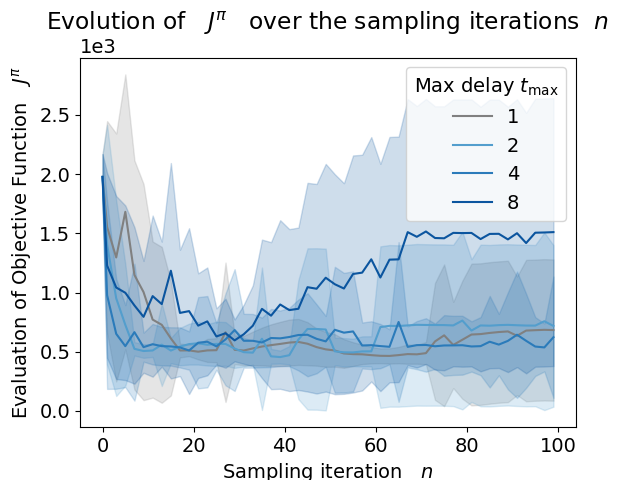}}
        \caption{Evolution of the objective function $\objectiveFunciton^{\MPCpolicyEstimator}$ to evaluate the system during training.}
        \label{fig:objective_function_evolution}
    \end{figure}

    First, the evolution of the amount of information gathered during sampling
    through a comparison of
    $(\expectedInformationGain_\samplingIteration^{\TIP})
    _{\samplingIteration=1}^{\samplingIteration_{\max}}$,
    and
    $(\expectedInformationGain_\samplingIteration^{\SMTIP})
    _{\samplingIteration=1}^{\samplingIteration_{\max}}$
    presented in
    Figure~\ref{fig:eig_evolution} to assess the impact of the SMDP extension.
    Second, the corresponding inter-decision times
    $(\randomInterdecisionTime_\samplingIteration)_
        {\samplingIteration=1}^{\samplingIteration_{\max}}$ are shown in
    Figure~\ref{fig:interdecision_time_evolution} to evaluate the necessity of
    temporal abstraction.
    Lastly, the evolution of the objective function
    $\objectiveFunciton^{\MPCpolicyEstimator}$ from 5 fixed initial conditions
    $\stateProcess_0$ is shown as a function of the sampling iteration
    $\samplingIteration$ in Figure~\ref{fig:objective_function_evolution} to
    analyse the effective results of the proposed method.
    For all the figures, the shaded area represents the standard error
    over the 10 independent trials.

    About the first point, one can observe that in all cases, the
    EIG is larger for SM-TIP than for TIP ($\timeDelayPoint_{\max} = 1$) until one-fourth of the sampling
    budget is reached.
    This suggests that the SMDP extension is beneficial to the information
    gathering process at the beginning of the sampling procedure.
    This may be explained by the fact that the inter-decision times allow
    to de-correlate the collected states via the same mechanism illustrated in
    Figure~\ref{fig:lorenz_autocorrelation_x3}.
    Note also that, in the case of Lorenz (Figure~\ref{fig:eig_evolution_lorenz}),
    the EIG after approximately half of the sampling procedure is superior
    for TIP than SM-TIP since more information
    (state-actions pairs minimising the mutual information) remain to be gathered.

    Examining the inter-decision times
    $(\randomInterdecisionTime_\samplingIteration)_
        {\samplingIteration=1}^{\samplingIteration_{\max}}$,
    it can first be observed that globally
    $\randomInterdecisionTime_\samplingIteration > 1$ for the SMDP algorithms
    (where $\timeDelayPoint_{\max} > 1$).
    This shows that the sequential maximal EIG is approximately reached for
    inter-decision times that are larger than the original MDP decision times.
    This confirms the relevance of temporal abstraction to increase the
    information gathering process.
    However, the inter-decision times are not necessarily always equal to
    $\timeDelayPoint_{\max}$, suggesting the more informative observations
    are not always the temporally most distant ones.

    Moving on to the objective function, in the case of the Lorenz system
    (Figure~\ref{fig:objective_function_evolution_lorenz}), the evaluation
    performances show the learning speed is greater for the SM-TIP settings
    ($\timeDelayPoint_{\max} > 1$)
    than for the TIP setting ($\timeDelayPoint_{\max} = 1$).
    For the Pendulum case
    (Figure~\ref{fig:objective_function_evolution_pendulum}),
    except for the SM-TIP setting where
    $\timeDelayPoint_{\max} = 8$, the proposed approach shows better sample
    complexity since very few iterations are required to reach optimality
    (light blue curves ($\timeDelayPoint_{\max} \in \{2, 4\}$)
    are below the grey curve ($\timeDelayPoint_{\max} = 1$) for the first
    (up to $\samplingIteration = 20$) sampling iterations.
    Furthermore, one of the reasons the $\timeDelayPoint_{\max} = 8$ fails
    to achieve optimal performances is likely the \emph{bootstrapping
    prediction error} (not shown in this document) which increases with
    $\timeDelayPoint_{\max}$.
    Indeed, as mentioned in Section~\ref{sec:method-experiments} due to the
    non-causal property of $\expectedInformationGain^{\SMTIP}$, there exists
    a trade-off between the temporal extension of the dynamics to reach the
    new region of the state space and the bootstrapping error which increases
    with the temporal extension.

    \section[Conclusion]{Conclusion}\label{sec:conclusion}


    This study demonstrates that, when restricted to the trajectory of the system, the total information gathered for a given sampling budget can be increased by introducing temporal abstraction through the usage of SMDPs.
    Results show that learning the dynamics of the Inverted Pendulum and the Lorenz system is more data-efficient with the use of temporally-extended actions.

    Future work may extend this methodology to more complex systems,
    leveraging the flexibility of SMDPs.
    These systems have the potential to reach highly informative regions and efficiently
    capture rapid changes in system dynamics, as the information content can be increased when considering the time resolution as a decision variable.

    In summary, this work offers a concise yet comprehensive glimpse into the
    potential of SMDPs in Model Predictive Control.
    The results on known systems establish a robust foundation for broader
    applications and unveil potential future advancements in control strategies.

    \acks{The authors acknowledge the support of the French Agence Nationale
    de la Recherche (ANR), under grant ANR-REASON (ANR-21-CE46-0008)\@.
    This work was performed using HPC resources from GENCI-IDRIS
        (Grant 2023-[AD011014278]).}

    \bibliography{l4dc2024_bibliography_post_review}

    \appendix

\end{document}